\documentclass[10pt,twocolumn,letterpaper]{article}

\usepackage{iccv}
\usepackage{times}
\usepackage{epsfig}
\usepackage{graphicx}
\usepackage{amsmath}
\usepackage{amssymb}
\usepackage{tabularx}
\usepackage{array}
\usepackage{booktabs}
\usepackage{multirow}
\usepackage{subfigure}

\usepackage[breaklinks=true,bookmarks=false,colorlinks=True, linkcolor=red]{hyperref}

\iccvfinalcopy 

\def\iccvPaperID{1727} 

\begin{document}

\title{CoupleNet: Coupling Global Structure with Local Parts for Object Detection}

\author{Yousong Zhu{$^{1,2}$} \quad Chaoyang Zhao{$^{1,2}$} \quad Jinqiao Wang{$^{1,2}$} \quad Xu Zhao{$^{1,2}$} \quad Yi Wu{$^{3,4}$} \quad Hanqing Lu{$^{1,2}$}\\
{$^1$}National Laboratory of Pattern Recognition, Institute of Automation, \\Chinese Academy of Sciences, Beijing, China\\
{$^2$}University of Chinese Academy of Sciences\\
{$^3$}School of Technology, Nanjing Audit University, Nanjing, China\\
{$^4$}Department of Medicine, Indiana University, Indianapolis, USA\\
{\tt\small \{yousong.zhu, chaoyang.zhao, jqwang, xu.zhao, luhq\}@nlpr.ia.ac.cn, ywu.china@gmail.com}
}


\maketitle

\begin{abstract}

  The region-based Convolutional Neural Network (CNN) detectors such as Faster R-CNN or R-FCN have already shown promising results for object detection by combining the region proposal subnetwork and the classification subnetwork together. Although R-FCN has achieved higher detection speed while keeping the detection performance, the global structure information is ignored by the position-sensitive score maps. To fully explore the local and global properties, in this paper, we propose a novel fully convolutional network, named as CoupleNet, to couple the global structure with local parts for object detection. Specifically, the object proposals obtained by the Region Proposal Network (RPN) are fed into the the coupling module which consists of two branches. One branch adopts the position-sensitive RoI (PSRoI) pooling to capture the local part information of the object, while the other employs the RoI pooling to encode the global and context information. Next, we design different coupling strategies and normalization ways to make full use of the complementary advantages between the global and local branches. Extensive experiments demonstrate the effectiveness of our approach. We achieve state-of-the-art results on all three challenging datasets, \ie a mAP of $82.7\%$ on VOC07, $80.4\%$ on VOC12, and $34.4\%$ on COCO.
  Codes will be made publicly available\footnote {\url{https://github.com/tshizys/CoupleNet}}.
\end{abstract}

\section{Introduction}

General object detection requires to accurately locate and classify all targets in the image or video. Compared to specific object detection, such as face, pedestrian and vehicle detection, general object detection often faces more challenges due to the large inter-class appearance differences. The variations arise not only from changes in a variety of non-rigid deformations, but also due to the truncations, occlusions and inter-class interference. However, no matter how complicated the objects are, when humans identify a target, the recognition of object categories is subserved by both a global process that retrieves structural information and a local process that is sensitive to individual parts. This motivates us to build a detection model that fused both global and local information.

\begin{figure}[t]
\begin{center}
\includegraphics[width=1.0\linewidth]{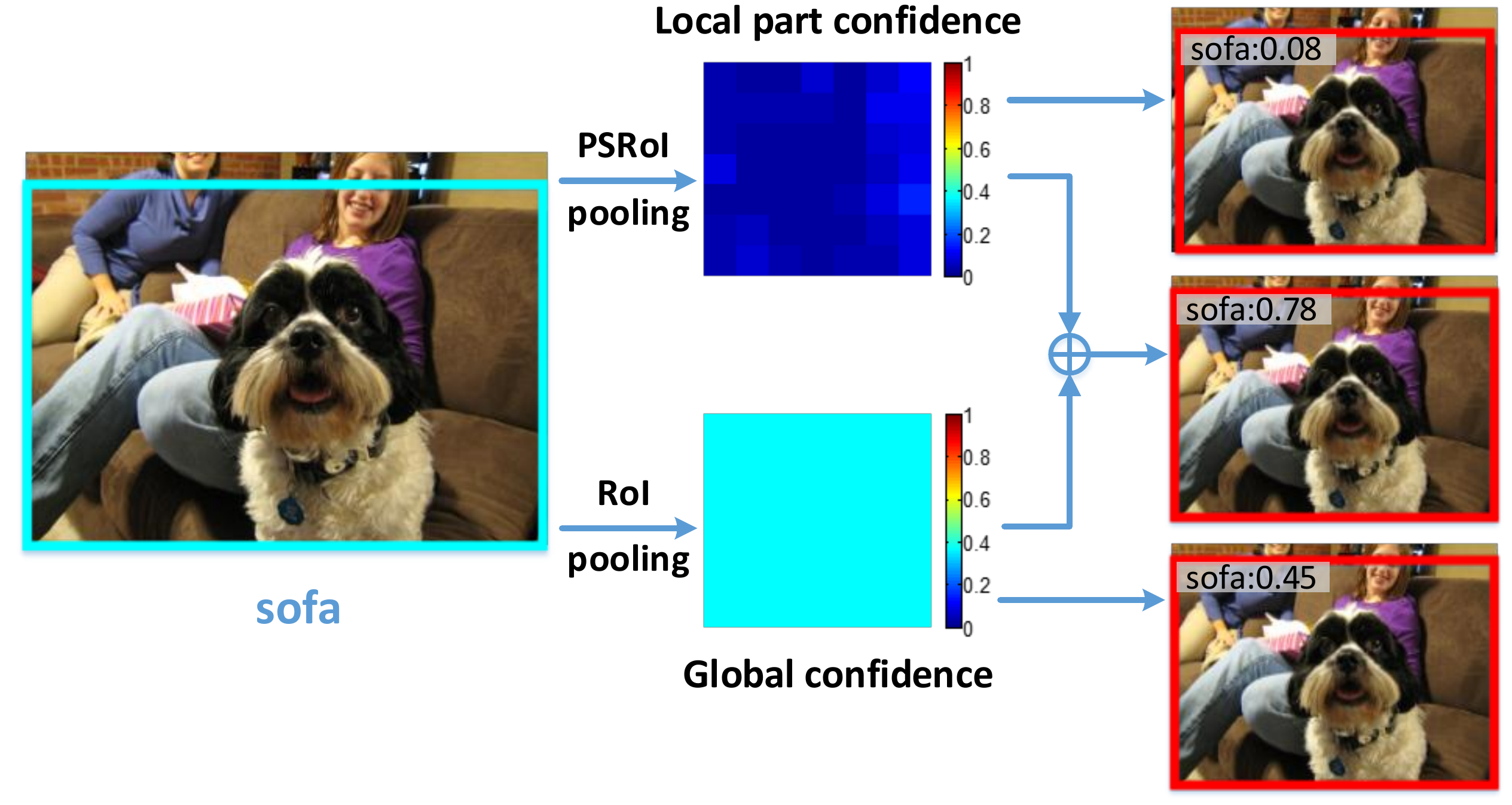}
\end{center}
  \caption{A toy example of object detection by combing local and global information. Only considering the local part information or global structure leads to low confidence score. By coupling the two kinds of information together, we can detect the sofa accurately with a confidence score of 0.78. Best viewed in color.}
\label{introduction}
\end{figure}

With the revival of Convolutional Neural Networks~\cite{krizhevsky2012imagenet} (CNN), CNN-based object detection pipelines~\cite{girshick2015fast, girshick2014rich, li2016r, ren2015faster} have been proposed consecutively and made impressive improvements in generic benchmarks, \eg PASCAL VOC~\cite{everingham2010pascal} and MS COCO~\cite{lin2014microsoft}. As two representative region-based CNN approaches, Fast/Faster R-CNN~\cite{girshick2015fast, ren2015faster} uses a certain subnetwork to predict the category of each region proposal while R-FCN~\cite{li2016r} conducts the inference with the position-sensitive score maps. Through removing the RoI-wise subnetwork, R-FCN  has achieved higher detection speed while keeping the detection performance. However, the global structure information is ignored by the PSRoI pooling. As shown in Figure~\ref{introduction}, using PSRoI pooling to extract local part information for final object category prediction, R-FCN leads to a low confidence score of 0.08 for the sofa detection since the local responses of sofa are disturbed by a women and a dog (they are also the categories that need to be detected). Conversely, the global structure of sofa could be extracted by the RoI pooling, but the confidence score is 0.45, which is also very low for the incomplete structure of sofa. By coupling the global confidence with the local part confidence together, we can obtain a more reliable prediction with the confidence score of 0.78.

In fact, the idea of fusing global and local information together is widely used in lots of visual tasks. In fingerprint recognition, Gu \etal~\cite{gu2006fingerprint} combined the global orientation field and local minutiae cue to largely improve the performance. In clique-graph matching, Nie \etal~\cite{nie2015clique} proposed a clique-graph matching method by preserving global clique-to-clique correspondence and local unary and pairwise correspondences. In scene parsing, Zhao \etal~\cite{zhao2016pyramid} designed a pyramid pooling module to effectively extract hierarchical global contextual prior, and then concatenated it with the local FCN feature to improve the performance. In traditional object detection, Felzenszwalb \etal~\cite{felzenszwalb2010object} incorporated a global root model and several finer local part models to represent highly variable objects. All of which show that effective combination of the global structural properties and local fine-grained details can achieve complementary advantages.

Therefore, to fully explore the global and local clues, in this paper, we propose a novel full convolutional network named as CoupleNet, to couple the global structure and local parts to boost the detection accuracy.
Specifically, the object proposals obtained by the RPN are fed into the coupling module which consists of two branches. One branch adopts the PSRoI pooling to capture the local part information of the object, while the other employs the RoI pooling to encode the global and context information. Moreover, we design different coupling strategies and normalization ways to make full use of the complementary advantages between the global and local branches. With the coupling structure, our network can jointly learn the local, global and context expression of the objects, which makes the model have a more powerful representation capacity and generalization ability. Extensive experiments demonstrate that CoupleNet can significantly improve the detection performance. Our detector shows competitive results on PASCAL VOC 07/12 and MS COCO compared to other state-of-the-art detectors, even with model ensemble approaches.

In summary, our main contributions are as follows:

1. We propose a unified fully convolutional network to jointly learn the local, global and context information for object detection.

2. We design different normalization methods and coupling strategies to mine the compatibility and complementarity between the global and local branches.

3. We achieve the state-of-the-art results on all three challenging datasets, \ie a mAP of $82.7\%$ on VOC07, $80.4\%$ on VOC12, and $34.4\%$ on MS COCO.

\section{Related work}
Before the arrival of CNN, visual tasks have been dominated by traditional paradigms~\cite{dollar2014fast, felzenszwalb2010object, viola2004robust, wang2014bilayer, wang2008multimodal}. As one of an outstanding framework, DPM~\cite{felzenszwalb2010object} described the object system using mixtures of multi-scale deformable part models, including a coarse global root model and several finer local part models. The root model extracts structural information of the objects, while the part models capture local appearance properties of an object. The sum of root response and weighted average response of each part is used as the final confidence of an object. Although DPM provides an elegant framework for object detection, the hand-crafted features, \ie improved HOG~\cite{dalal2005histograms}, are not discriminative enough to express the diversity of object categories. This is also the main reason that CNN completely surpassed the traditional methods in a short period time.

In order to leverage the great success of deep neural networks for image classification~\cite{he2016deep, krizhevsky2012imagenet}, considerable object detection methods based on deep learning have been proposed~\cite{girshick2014rich, he2014spatial, liu2016ssd, redmon2016you, zhu2016scale}. Although there are end-to-end detection frameworks, like SSD~\cite{liu2016ssd}, YOLO~\cite{redmon2016you} and DenseBox~\cite{huang2015densebox}, region-based systems (\ie Fast/Faster R-CNN~\cite{girshick2015fast, ren2015faster} and R-FCN~\cite{li2016r}) still dominate the detection accuracy on generic benchmarks~\cite{everingham2010pascal, lin2014microsoft}.

Compared to the end-to-end framework, the region-based systems have several advantages. Firstly, by exploiting a divide-and-conquer strategy, the two-step framework is more stable and easier to converge. Secondly, without the complicated data augmentation and training skills, you can still easily achieve state-of-the-art performance. The main reason for these advantages is that there is a certain structure~\cite{girshick2015fast, li2016r, ren2015faster} to encode translation variance features for each proposal, since in deep networks, higher-layers contain more semantic meaning and less location information. As a consequence, a RoI-wise subnetwork~\cite{girshick2015fast, ren2015faster} or a position-sensitive RoI pooling layer~\cite{li2016r} is used to achieve the translation variance in region-based systems. However, all the existing region-based systems utilize either the region-level or part-level features to learn the variations, where each one alone is not representative enough for a variety of challenging situations. Therefore, this motivates us to design a certain structure to take advantages of both the global and local features.

\begin{figure*}[thbp]
\begin{center}
\includegraphics[width=0.95\linewidth]{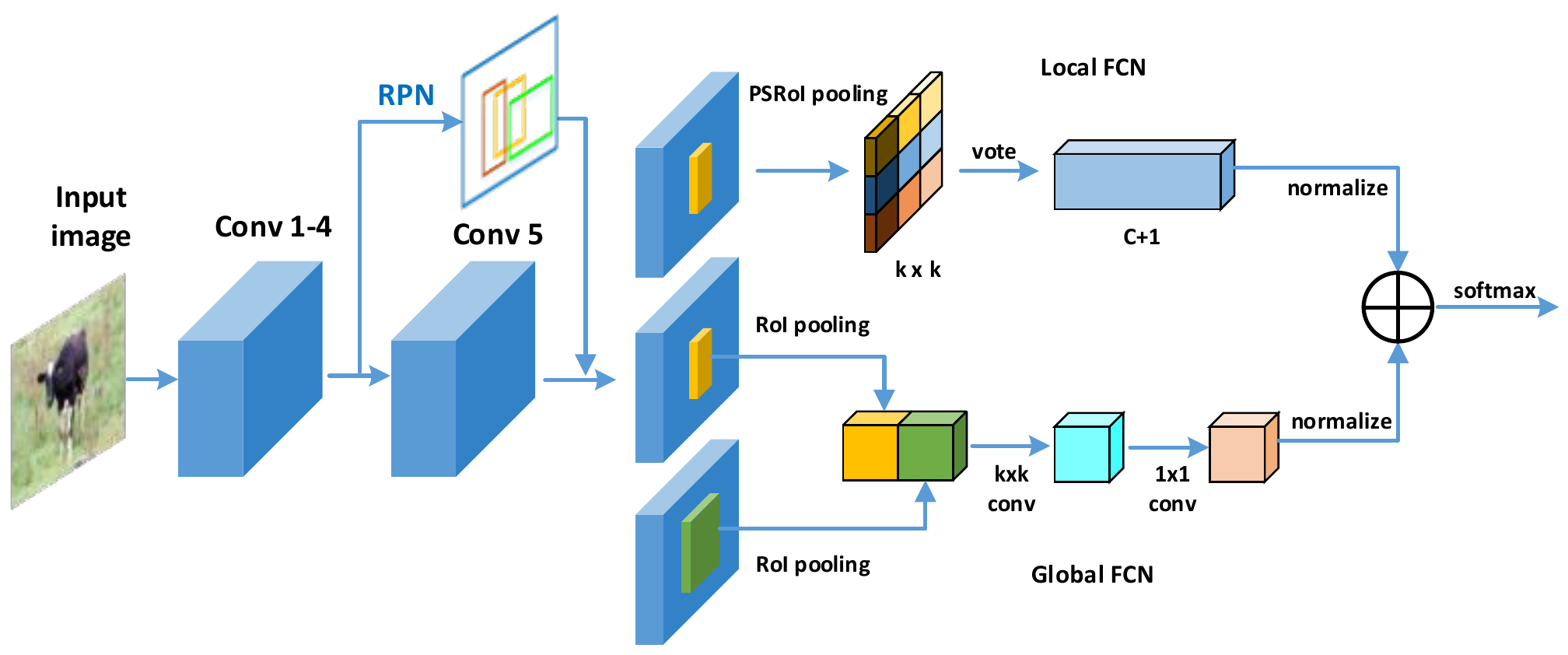}
\end{center}
  \caption{The architecture of the proposed CoupleNet. We use ResNet-101 as the basic feature extraction network. Given an input image, we first exploit Region Proposal Network (RPN)~\cite{ren2015faster} to generate candidate proposals. Then each proposal flows to two different branches: local FCN and global FCN, in order to extract the global structure information and learn the object-specific parts respectively. Finally the output of the two branches are coupled together to predict the object categories.}
\label{framework}
\end{figure*}

In addition, context~\cite{wang2014spatiotemporal} is known to play an important role in visual recognition. Considerable works have been proposed for exploting context in object detection. Bell \etal~\cite{bell16ion} explored the use of recurrent neural networks to model the contextual information. Gidaris \etal~\cite{gidaris2015object} proposed to utilize multiple contextual regions around the object. Cai \etal~\cite{cai2016unified} collected the context by padding the proposals for pedestrian and car detection. Similar to these works, we also absorb the context prior to enhance the global feature representation.

\section{CoupleNet}
In this section, we first introduce the architecture of the proposed CoupleNet for object detection. Then we explain in detail how we incorporate local representations, global appearance and contextual information for robust object detection.

\subsection{Network architecture}
The architecture of our proposed CoupleNet is illustrated in Figure~\ref{framework}. Our CoupleNet includes two different branches: a) a local part-sensitive fully convolutional network to learn the object-specific parts, denoted as local FCN; b) a global region-sensitive fully convolutional network to encode the whole appearance structure and context prior of the object, denoted as global FCN. We first use the ImageNet pre-trained ResNet-101 released in~\cite{he2016deep} to initialize our network. For our detection task, we remove the last average pooling layer and the \textit{fc} layer. Given an input image, we extract candidate proposals by using the Region Proposal Network (RPN), which also shares convolution features with CoupleNet following~\cite{ren2015faster}. Then each proposal flows to two different branches: the local FCN and the global FCN. Finally, the output of global and local FCN are coupled together as the final score of the object. We also perform class-agnostic bounding box regression in a similar way.

\subsection{Local FCN}
\begin{figure}[!t]
\begin{center}
\includegraphics[width=0.9\linewidth]{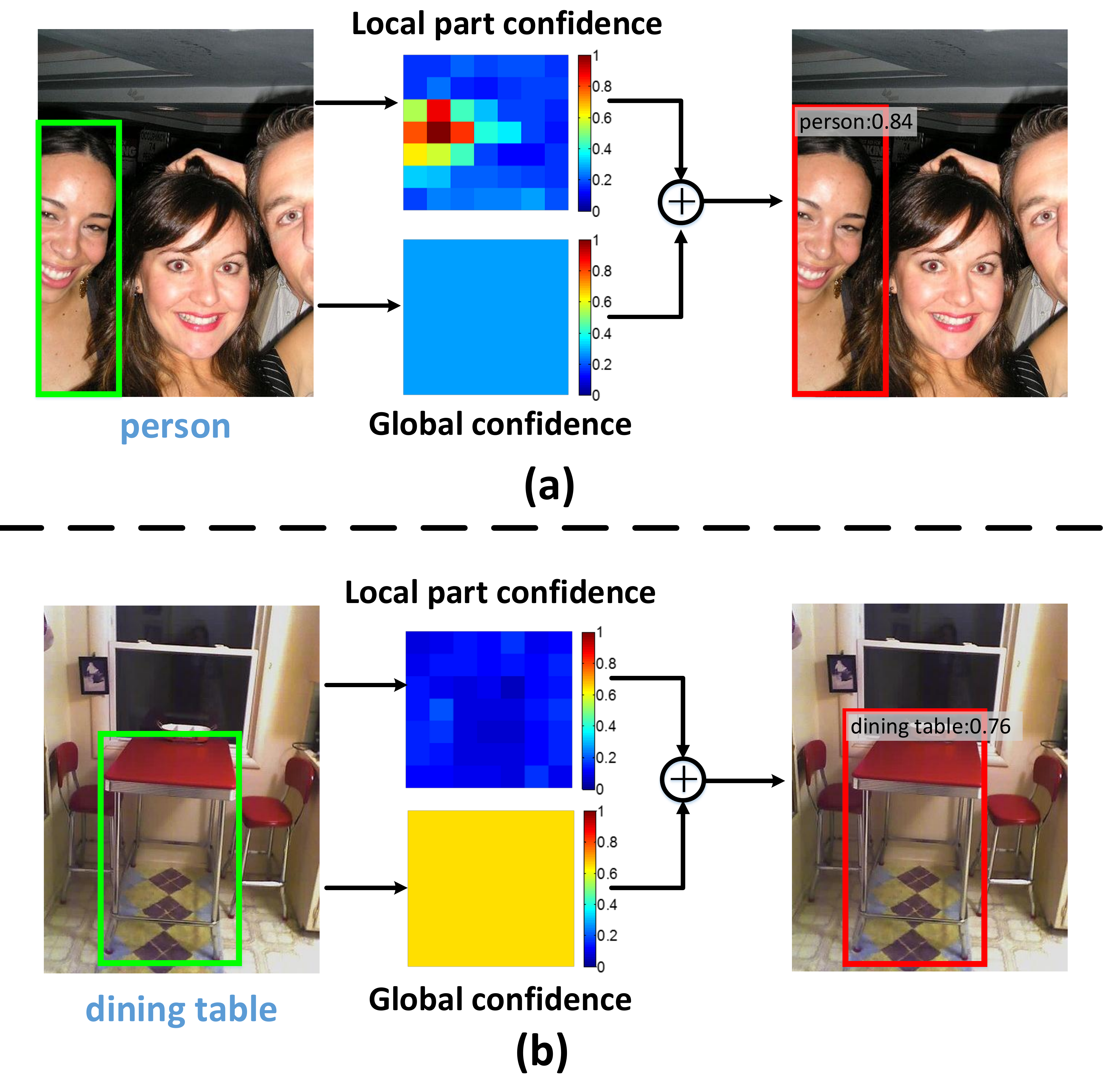}
\end{center}
  \caption{An intuitive description of CoupleNet for object detection. (a) It is difficult to determine the target by using the global structure information alone for objects with truncations. (b) Moreover,
  for those having simple spatial structure and encompassing considerable background in the bounding box, \eg dining table, it is also not enough to use local parts alone to make robust predictions. Therefore, an intuitive idea is to simultaneously couple global structure with local parts to effectively boost the confidence. Best viewed in color.}
\label{overview}
\end{figure}

To effectively capture the specific fine-grained parts in local FCN, we construct a set of part-sensitive score maps by appending a 1x1 convolutional layer with $k^2(C+1)$ channels, where $k$ means we divide the object into $k\times k$ local parts (here $k$ is set to the default value 7) and $C+1$ is the number of object categories plus background. For each category, there are totally $k^2$ channels and each channel is responsible for encoding a specific part of the object. The final score of a category is determined by voting the $k^2$ responses. Here we use position-sensitive RoI pooling layer in~\cite{li2016r} to extract object-specific parts and we simply perform average pooling for voting. Then, we obtain a $(C+1)$-d vector which indicates the probability that the object belongs to each class. This procedure is equivalent to dividing a strong object category decision into the sum of multiple weak classifiers, which serves as the ensemble of several part models. Here we refer this part ensemble as local structure representation. As shown in Figure~\ref{overview}(a), for the truncated person, one can hardly get a strong response from the global description of the person due to truncation, on the contrary, our local FCN can effectively capture several specific parts, such as human nose, mouth, \etc, which correspond to the regions with large responses in the feature map. We argue that the local FCN is much concerned with the internal structure and components, which can effectively reflect the local properties of visual object, especially when the object is occluded or the whole boundary is incomplete. However, for those having simple spatial structure and encompassing considerable background in the bounding box, \eg dining table, the local FCN alone is difficult to make robust predictions. Thus it is necessary to add the global structure information to enhance the discrimination.

\subsection{Global FCN}
For the global FCN, we aim to describe the object by using the whole region-level features. Firstly, we attach a 1024-d 1x1 convolutional layer after the last convolutional block in ResNet-101 for reducing the dimension. Due to the diverse size of the object, we insert a RoI pooling layer in~\cite{girshick2015fast} to extract a fixed-length feature vector as the global structure description of the object. Secondly, we use two convolutional layers with kernal size $k\times k$ and $1\times 1$ respectively ($k$ is set to the default value 7) to further abstract the global representation of RoI. Finally, the output of 1x1 convolution is fed into the classifier whose output is also a $(C+1)$-d vector.

In addition, context prior is the most basic and important factor for visual recognition tasks. For example, the boat usually travels in the water while is unlikely to fly in the sky. Despite the higher layers in deep neural network can involve the spatial context information around the objects due to the large receptive field, Zhou \etal~\cite{zhou2014object} have shown that the practical receptive field is actually much smaller than the theoretical one. Therefore, it is necessary to explicitly collect the surrounding information to reduce the chance of misclassification. To enhance the feature representation ability of the global FCN, here we introduce the contextual information as an effective supplement. Specifically, we extend the context region by 2 times larger than the size of original proposal. Then the features RoI pooled from the original region and context region are concatenated together and fed into the latter RoI-wise subnetwork.As shown in Figure~\ref{framework}, the context region is embedded into the global branch to extract a more complete appearance structure and discriminative prior representation, which will help the classifier to better identity the object categories.

Due to the RoI pooling operation, the global FCN describes the proposal as a whole with CNN features, which can be seen as a global structure description of the object. Therefore, it can easily deal with the objects with intact structure and finer scale. As shown in Figure \ref{overview}(b), our global FCN shows a large confidence for the dining table. However, in most cases, natural scenes consist of considerable objects with occlusions or truncations, making the detection more difficult. Figure~\ref{overview}(a) shows that using the global structure information alone can hardly make a confident prediction for the truncated person. By adding local part structural supports, the detection performance can be significantly boosted. Therefore, it is essential to combine both local and global descriptions for a robust detection.

\subsection{Coupling structure}
To match the same order of magnitude, we apply a normalization operation to the output of local and global FCN before they are combined together. We explored two different methods to perform normalization: an L2 normalization layer or a 1x1 convolutional layer to model the scale. Meanwhile, how to couple the local and global output is also a problem that needs to be researched. Here, we investigated three different coupling methods: element-wise sum, element-wise product and element-wise maximum. Our experiments show that using 1x1 convolution along with element-wise sum achieves the best performance and we will discuss it in Section \ref{Ablation}.

With the coupling structure, CoupleNet simultaneously exploits the local parts, global structure and context prior for object detection. The whole network is fully convolutional and benefits from approximate joint training and multi-task learning. We also note that the global branch can be regarded as a lightweight Faster R-CNN, in which all learnable parameters are from convolutional layers and the depth of RoI-wise subnetwork is only two. Therefore, the computational complexity is far less than the subnetwork in ResNet-based Faster R-CNN system whose depth is ten. As a consequence, our CoupleNet can perform the inference efficiently, which runs slightly slower than R-FCN but much more faster than Faster R-CNN.
\section{Experiments}
We train and evaluate our method on three challenging object detection datasets: PASCAL VOC2007, VOC2012 and MS COCO. Since all these three datasets contain a variety of circumstances, which can sufficiently verify the effectiveness of our method. We demonstrate state-of-the-art results on all three datasets without bells and whistles.
\subsection{Ablation studies on VOC2007}\label{Ablation}
We first perform experiments on PASCAL VOC 2007 with 20 object categories for detailed analysis of our proposed CoupleNet detector. We train the models on the union set of VOC 2007 trainval and VOC 2012 trainval (``07+12") following~\cite{ren2015faster}, and evaluate on VOC 2007 test set. Object detection accuracy is measured by mean Average Precision (mAP), all the ablation experiments use \textit{single-scale} training and testing, and we did not add the context prior.

\textbf{Normalization.} Since features extracted form different layers of CNN show various of scales, it is essential to normalize different features before coupling them together. Bell \etal~\cite{bell16ion} proposed to use L2 normalization to each RoI-pooled feature and re-scale back up by a empirical scale, which shows a great gain on VOC dataset. In this paper, we also explore two different normalization ways to normalize the output of local and global FCN: an L2 normalization layer or a 1x1 convolutional layer to learn the scale.

As shown in Table~\ref{coupling strategies}, we find that the use of L2 normalization decreases the performance greatly, even worse than the direct addition (without any normalization ways). To explain such a phenomenon, we measured the outputs of two branches before and after L2 normalization. We further found that L2 normalization reduces the output gap between different categories, which results in a smaller score gap. As we know, a small score gap between different categories always means the classifier can not make a confident prediction. Therefore, we assume that this is the reason for the performance degradation. Moreover, we also exploit a 1x1 convolution to adaptively learn the scales between the global and local branches. Table~\ref{coupling strategies} shows that using 1x1 convolution increases by $0.6$ points compared to the direct addition and $2.2$ points over R-FCN. Therefore, we use 1x1 convolution to replace the L2 normalization in the following experiments.

\begin{table}[htbp]
\begin{center}
\scalebox{0.9}{
\begin{tabular}{cccc}
\toprule
Normalization methods & SUM & PROD & MAX \\
\midrule\noalign{\smallskip}
eltwise & \textbf{81.1}  & - & 80.7 \\
\noalign{\smallskip}
L2+eltwise & \textbf{80.3} & 63.5 & 78.2 \\
\noalign{\smallskip}
1x1 conv+eltwise & \textbf{81.7}  & - & 81.3 \\
\bottomrule
\end{tabular}}
\end{center}
\caption{\textbf{Effects of different normalization operation and coupling methods.} Metric: detection mAP($\%$) on VOC07 test. eltwise: combine the output from global and local FCN directly. L2+eltwise: use L2 normalization to normalize the output. 1x1 conv+eltwise: use 1x1 convolution to learn the scale.}
\label{coupling strategies}
\end{table}

\textbf{Coupling strategy.} We explore three different response coupling strategies: element-wise sum, element-wise product and element-wise maximum. Table~\ref{coupling strategies} shows the comparison results for the above three different implementations. We can see that the element-wise sum always achieves the best performance even though in different normalization methods. Generally, current advanced residual networks~\cite{he2016deep} also use element-wise sum as the effective way to integrate information from previous layers, which greatly facilitates the circulation of information and achieves the complementary advantages. For element-wise product, we argue that the system is relatively unstable and is susceptible to the weak side, which results in a large gradient to update the weak branch that makes it difficult to converge. For element-wise maximum, it equals to an ensemble model within the network to some extent, which losts the advantages of mutual support compared to element-wise sum when both two branches are failed to detect the object. Moreover, a better coupling strategy can be taken into consideration as the future work to further improve the accuracy, such as designing a more subtle nonlinear structure to learn the coupling relationship.

\begin{table*}[thbp]
\begin{center}
\scalebox{0.95}{
\begin{tabular}{c|c|c|c|c}
\hline
 & training data & mAP ($\%$) on VOC07 & GPU & test time (ms/img) \\
 \hline
Faster R-CNN~\cite{he2016deep} & 07+12 & 76.4 & K40 & 420 \\
\hline
R-FCN~\cite{li2016r} & 07+12 & 79.5 & TITAN X & 83 \\
R-FCN \begin{small}\textit{multi-sc train}\end{small}~\cite{li2016r} & 07+12 & 80.5 & TITAN X & 83 \\
\hline
CoupleNet & 07+12 & \textbf{81.7} & TITAN X & 102 \\
CoupleNet \begin{footnotesize}\textit{context}\end{footnotesize} & 07+12 & \textbf{82.1} & TITAN X & 122 \\
CoupleNet \begin{footnotesize}\textit{context multi-sc train}\end{footnotesize} & 07+12 & \textbf{82.7} & TITAN X & 122 \\
\hline
\end{tabular}}
\end{center}
\caption{\textbf{Comparisons with Faster R-CNN and R-FCN using ResNet-101.} 128 samples are used for backpropagation and the top 300 proposals are selected for testing following~\cite{li2016r}. The input resolution is 600x1000. We also note that the TITAN X used here is the new Pascal architecture along with CUDA 8.0 and cuDNN-v5.1. ``07+12": VOC07 trainval union with VOC12 trainval. \textit{context}: add the context prior to assist the global branch.}
\label{comparions with Faster and RFCN}
\end{table*}

\begin{table*}[htbp]
\begin{center}
\begin{tabularx}{\linewidth}{>{\footnotesize}p{1.98cm}<{\centering}|>{\scriptsize}p{0.95cm}<{\centering}|>{\small}p{0.48cm}<{\centering}|>{\footnotesize}X<{\centering}>{\footnotesize}X<{\centering}>{\footnotesize}X<{\centering}>{\footnotesize}X<{\centering}>{\footnotesize}X<{\centering}
>{\footnotesize}X<{\centering}>{\footnotesize}X<{\centering}>{\footnotesize}X<{\centering}>{\footnotesize}X<{\centering}>{\footnotesize}X<{\centering}>{\footnotesize}X<{\centering}>{\footnotesize}X<{\centering}>{\footnotesize}X<{\centering}>{\footnotesize}X<{\centering}>{\footnotesize}X<{\centering}>{\footnotesize}X<{\centering}>{\footnotesize}X<{\centering}
>{\footnotesize}X<{\centering}>{\footnotesize}X<{\centering}>{\footnotesize}X<{\centering}}
\hline
Method&Train&\scriptsize{mAP}&\scriptsize{aero}&\scriptsize{bike}&\scriptsize{bird}&\scriptsize{boat}&\scriptsize{bottle}&\scriptsize{bus}&\scriptsize{car}&\scriptsize{cat}&\scriptsize{chair}&\scriptsize{cow}&\scriptsize{table}&\scriptsize{dog}&
\scriptsize{horse}&\scriptsize{mbike}&\scriptsize{persn}&\scriptsize{plant}&\scriptsize{sheep}&\scriptsize{sofa}&\scriptsize{train}&\scriptsize{tv} \\
\hline
ION~\cite{bell16ion} & 07+12+S & 76.5 & 79.2 & 79.2 & 77.4 & 69.8 & 55.7 & 85.2 & 84.2 & \textbf{89.8} & 57.5 & 78.5 & 73.8 & 87.8 & 85.9 & 81.3 & 75.3 & 49.7 & 76.9 & 74.6 & 85.2 & \textbf{82.1} \\
HyperNet~\cite{kong2016hypernet} & 07+12 & 76.3 & 77.4 & 83.3 & 75.0 & 69.1 & 62.4 & 83.1 & 87.4 & 87.4 & 57.1 & 79.8 & 71.4 & 85.1 & 85.1 & 80.0 & 79.1 & 51.2 & 79.1 & 75.7 & 80.9 & 76.5\\
SSD300$^*$~\cite{liu2016ssd} & 07+12 & 77.5 & 79.5 & 83.9 & 76.0 & 69.6 & 50.5 & 87.0 & 85.7 & 88.1 & 60.3 & 81.5 & \textbf{77.0} & 86.1 & 87.5 & 83.9 & 79.4 & 52.3 & 77.9 & 79.5 & 87.6 & 76.8 \\
SSD512$^*$~\cite{liu2016ssd} & 07+12 & 79.5 & 84.8 & 85.1 & 81.5 & 73.0 & 57.8 & 87.8 & 88.3 & 87.4 & 63.5 & 85.4 & 73.2 & 86.2 & 86.7 & 83.9 & 82.5 & 55.6 & 81.7 & 79.0 & 86.6 & 80.0 \\
Faster$^\S$~\cite{he2016deep} & 07+12 & 76.4 & 79.8 & 80.7 & 76.2 & 68.3 & 55.9 & 85.1 & 85.3 & \textbf{89.8} & 56.7 & 87.8 & 69.4 & 88.3 & 88.9 & 80.9 & 78.4 & 41.7 & 78.6 & 79.8 & 85.3 & 72.0 \\
R-FCN~\cite{li2016r} & 07+12 & 80.5 & 79.9 & \textbf{87.2} & 81.5 & 72.0 & 69.8 & 86.8 & 88.5 & 89.8 & 67.0 & \textbf{88.1} & 74.5 & \textbf{89.8} & \textbf{90.6} & 79.9 & 81.2 & 53.7 & 81.8 & 81.5 & 85.9 & 79.9 \\
CoupleNet [ours] & 07+12 & \textbf{82.7} & \textbf{85.7} & 87.0 & \textbf{84.8} & \textbf{75.5} & \textbf{73.3} & \textbf{88.8} & \textbf{89.2} & 89.6 & \textbf{69.8} & 87.5 & 76.1 & 88.9 & 89.0 & \textbf{87.2} & \textbf{86.2} & \textbf{59.1} & \textbf{83.6} & \textbf{83.4} & \textbf{87.6} & 80.7 \\
\hline
\end{tabularx}
\end{center}
\caption{\textbf{Results on PASCAL VOC 2007 test set.} The first four methods use VGG16 and the latter three use ResNet-101 as the base network. For fair comparison, we only list the results of single model without multi-scale testing, ensemble or iterative box regression tricks in testing phase. ``07+12": VOC07 trainval union with VOC12 trainval. ``07+12+S": VOC07 trainval union with VOC12 trainval plus segmentation labels. *: the results are updated using the latest models. \S: this entry is directly obtained from~\cite{he2016deep} without using OHEM.}
\label{comparisons with state-of-the-art}
\end{table*}
\textbf{Model ensemble.} Model ensemble is commonly used to improve the final detection performance, since diverse initialization of parameters and the randomness of training samples both lead to different performance for the same model. Although the differences and complementarities will be more pronounced for different models, the promotion is often very limited. As shown in Table~\ref{Model ensembling}, we also compare our CoupleNet with the model ensemble. For a fair comparison, we first re-implemented Faster R-CNN~\cite{he2016deep} using ResNet-101 and online hard example mining (OHEM)~\cite{shrivastava2016training}, which achieves a mAP of $79.0\%$ on VOC07 ($76.4\%$ in original paper without OHEM). We also re-implemented R-FCN with appropriate joint training using the public available code py-R-FCN\footnote {\url{https://github.com/Orpine/py-R-FCN}}, which achieves a slightly lower result compared to~\cite{li2016r} ($78.6\%$ vs. $79.5\%$). We use our reimplementation models to conduct the comparisons for consistency. We found that the promotion brought by model ensemble is less than 1 point. As shown in Table~\ref{Model ensembling}, it is far less than our method ($81.7\%$).

\begin{table}[htb]
\begin{center}
\scalebox{0.9}{
\begin{tabular}{c|c}
\hline
Method & mAP($\%$) \\
\hline
Faster-\small{\textit{ReIm}} & 79.0 \\
R-FCN-\small{\textit{ReIm}} & 78.6 \\
Global FCN & 78.5 \\
Faster\&R-FCN ensemble & 79.6 \\
Global FCN\&R-FCN ensemble & 79.4 \\
CoupleNet & \textbf{81.7} \\
\hline
\end{tabular}}
\end{center}
\caption{\textbf{CoupleNet \textit{vs.} model ensemble.} \textit{ReIm}: our reimplementation using OHEM. Global FCN: only the global branch of our network.}
\label{Model ensembling}
\end{table}

On the one hand, we argue that the naive model ensemble just combines the results together and does not essentially guide the learning process of the network, while our CoupleNet can simultaneously utilize the global and local information to update the network and to infer the final results. On the other hand, our method enjoys end-to-end training and there is no need to train multiple models, thus greatly reducing the training time.

\begin{table*}[!thbp]
\begin{center}
\begin{tabularx}{\linewidth}{>{\footnotesize}p{1.98cm}<{\centering}|>{\scriptsize}p{0.95cm}<{\centering}|>{\small}p{0.48cm}<{\centering}|>{\footnotesize}X<{\centering}>{\footnotesize}X<{\centering}>{\footnotesize}X<{\centering}>{\footnotesize}X<{\centering}>{\footnotesize}X<{\centering}
>{\footnotesize}X<{\centering}>{\footnotesize}X<{\centering}>{\footnotesize}X<{\centering}>{\footnotesize}X<{\centering}>{\footnotesize}X<{\centering}>{\footnotesize}X<{\centering}>{\footnotesize}X<{\centering}>{\footnotesize}X<{\centering}>{\footnotesize}X<{\centering}>{\footnotesize}X<{\centering}>{\footnotesize}X<{\centering}>{\footnotesize}X<{\centering}
>{\footnotesize}X<{\centering}>{\footnotesize}X<{\centering}>{\footnotesize}X<{\centering}}
\hline
Method&Train&\scriptsize{mAP}&\scriptsize{aero}&\scriptsize{bike}&\scriptsize{bird}&\scriptsize{boat}&\scriptsize{bottle}&\scriptsize{bus}&\scriptsize{car}&\scriptsize{cat}&\scriptsize{chair}&\scriptsize{cow}&\scriptsize{table}&\scriptsize{dog}&
\scriptsize{horse}&\scriptsize{mbike}&\scriptsize{persn}&\scriptsize{plant}&\scriptsize{sheep}&\scriptsize{sofa}&\scriptsize{train}&\scriptsize{tv} \\
\hline
ION~\cite{bell16ion} & 07+12+S & 76.4 & 87.5 & 84.7 & 76.8 & 63.8 & 58.3 & 82.6 & 79.0 & 90.9 & 57.8 & 82.0 & 64.7 & 88.9 & 86.5 & 84.7 & 82.3 & 51.4 & 78.2 & 69.2 & 85.2 & 73.5 \\
HyperNet~\cite{kong2016hypernet} & 07++12 & 71.4 & 84.2 & 78.5 & 73.6 & 55.6 & 53.7 & 78.7 & 79.8 & 87.7 & 49.6 & 74.9 & 52.1 & 86.0 & 81.7 & 83.3 & 81.8 & 48.6 & 73.5 & 59.4 & 79.9 & 65.7\\
SSD300$^*$~\cite{liu2016ssd} & 07++12 & 75.8 & 88.1 & 82.9 & 74.4 & 61.9 & 47.6 & 82.7 & 78.8 & 91.5 & 58.1 & 80.0 & 64.1 & 89.4 & 85.7 & 85.5 & 82.6 & 50.2 & 79.8 & \textbf{73.6} & 86.6 & 72.1 \\
SSD512$^*$~\cite{liu2016ssd} & 07++12 & 78.5 & \textbf{90.0} & 85.3 & 77.7 & 64.3 & 58.5 & \textbf{85.1} & \textbf{84.3} & 92.6 & 61.3 & 83.4 & 65.1 & 89.9 & 88.5 & \textbf{88.2} & 85.5 & 54.4 & 82.4 & 70.7 & 87.1 & \textbf{75.6} \\
Faster$^\S$~\cite{he2016deep} & 07++12 & 73.8 & 86.5 & 81.6 & 77.2 & 58.0 & 51.0 & 78.6 & 76.6 & 93.2 & 48.6 & 80.4 & 59.0 & 92.1 & 85.3 & 84.8 & 80.7 & 48.1 & 77.3 & 66.5 & 84.7 & 65.6 \\
R-FCN~\cite{li2016r} & 07++12 & 77.6 & 86.9 & 83.4 & 81.5 & 63.8 & 62.4 & 81.6 & 81.1 & 93.1 & 58.0 & 83.8 & 60.8 & \textbf{92.7} & 86.0 & 84.6 & 84.4 & 59.0 & 80.8 & 68.6 & 86.1 & 72.9 \\
CoupleNet [ours] & 07++12 & \textbf{80.4}$^\dagger$ & 89.1 & \textbf{86.7} & \textbf{81.6} & \textbf{71.0} & \textbf{64.4} & 83.7 & 83.7 & \textbf{94.0} & \textbf{62.2} & \textbf{84.6} & \textbf{65.6} & \textbf{92.7} & \textbf{89.1} & 87.3 & \textbf{87.7} & \textbf{64.3} & \textbf{84.1} & 72.5 & \textbf{88.4} & 75.3 \\
\hline
\end{tabularx}
\end{center}
\caption{\textbf{Results on PASCAL VOC 2012 test set.} For fair comparison, we only list the results of single model without multi-scale testing, ensemble or iterative box regression tricks in testing phase. ``07++12": the union set of VOC07 trainval+test and VOC12 trainval. ``07+12+S": VOC07 trainval union with VOC12 trainval plus segmentation labels. *: results are updated using the latest models. \S: this entry is directly obtained from~\cite{he2016deep} without using OHEM. $\dagger$: \url{http://host.robots.ox.ac.uk:8080/anonymous/M5CQTL.html}.}
\label{VOC12}
\end{table*}

\begin{table*}[!htbp]
\begin{center}
\begin{tabularx}{\linewidth}{>{\footnotesize}p{2.75cm}|>{\footnotesize}p{1cm}<{\centering}|>{\small}p{0.45cm}<{\centering}|>{\small}p{0.55cm}<{\centering}|>{\small}p{0.7cm}<{\centering}|>{\small}p{0.65cm}<{\centering}
>{\small}p{0.95cm}<{\centering}>{\small}p{0.5cm}<{\centering}|>{\small}p{0.5cm}<{\centering}>{\small}p{0.6cm}<{\centering}>{\small}p{0.9cm}<{\centering}|>{\small}p{0.65cm}<{\centering}>{\small}p{0.95cm}<{\centering}>{\small}p{0.4cm}<{\centering}}
\hline
Method& train data&\footnotesize{AP}&\footnotesize{AP @0.5}&\footnotesize{AP @0.75}&\footnotesize{AP small}&\footnotesize{AP medium}&\footnotesize{AP large}&\footnotesize{AR max=1}&\footnotesize{AR max=10}&\footnotesize{AR max=100}&\footnotesize{AR small}&\footnotesize{AR medium}&\footnotesize{AR large} \\
\hline
SSD300$^*$~\cite{liu2016ssd} & \scriptsize{trainval35k}&25.1&43.1&25.8&6.6&25.9&41.4&23.7&35.1&37.2&11.2&40.4&58.4 \\
SSD512$^*$~\cite{liu2016ssd} & \scriptsize{trainval35k}&28.8&48.5&30.3&10.9&31.8&43.5&26.1&39.5&42.0&16.5&46.6&60.8 \\
\hline
ION~\cite{bell16ion} & \scriptsize{train+S}&24.9&44.7&25.3&7.0&26.1&40.1&23.9&33.5&34.1&10.7&38.8&54.1 \\
Faster+++~\cite{he2016deep} & \scriptsize{trainval}&\textbf{34.9} &\textbf{55.7} &- &15.6 &38.7 &50.9 &- &- &- &-  &- &- \\
R-FCN~\cite{li2016r} & \scriptsize{trainval}  &29.2 &51.5 &- &10.3 &32.4 &43.3 &- &- &- &-  &- &- \\
R-FCN \begin{scriptsize}\textit{multi-sc train}\end{scriptsize}~\cite{li2016r} & \scriptsize{trainval}  &29.9 &51.9 &- &10.8 &32.8 &45.0 &- &- &- &-  &- &- \\
\hline
CoupleNet & \scriptsize{trainval}  &\textbf{33.1} &\textbf{53.5} &35.4 &11.6 &36.3 &50.1 &29.3 &43.8 &45.2 &18.7 &51.4 &67.9 \\
CoupleNet \begin{scriptsize}\textit{multi-sc train}\end{scriptsize} & \scriptsize{trainval}  &\textbf{34.4} &\textbf{54.8} &37.2 &13.4 &38.1 &50.8 &30.0 &45.0 &46.4 &20.7  &53.1 &68.5 \\
\hline
\end{tabularx}
\end{center}
\caption{\textbf{Results on COCO 2015 test-dev.} The COCO metric AP is evaluated at IoU thresholds ranging from 0.5 to 0.95. AP@0.5: PASCAL-type metric, IoU=0.5. AP@0.75: evaluate at IoU=0.75. ``train+S": train set plus segmentation labels.}
\label{COCO}
\end{table*}


\textbf{Amount of parameters.} Since our CoupleNet introduces a few more parameters compared with the single branch detectors, to further verify effectiveness of the coupling structure, here we increase the parameters of the prediction head for each single branch implementation to maintain the same amount of parameters with CoupleNet for comparison. In detail, we add a new residual variant block with three convolution layers, where the kernel size is 1x1x256, 3x3x256 and 1x1x1024 respectively, to the prediction sub-network. We found that the standard R-FCN with one or two extra heads got a mAP of $78.8\%$ and $78.7\%$ respectively in VOC07, which is slightly higher than our re-implemented version ($78.6\%$) in ~\cite{li2016r} as shown in Table~\ref{Model ensembling}. Meanwhile, our global FCN, which performs the ROI Pooling on top of conv5, got a relative higher gain (a mAP of $79.3\%$ for one head, $79.0\%$ for two heads). The results indicate that simply adding more prediction layers obtains a very limited performance gain, while our coupling structure shows more discriminative power with the same amount of parameters.

\subsection{Results on VOC2007}
Using the public available ResNet-101 as the initialization model, we note that our method is easy to follow and the hyper-parameters for training are the same as in~\cite{li2016r}. Similarly, we use the dilation strategy to reduce the effective stride of ResNet-101, just as~\cite{li2016r} shows, thus both the global and local branches have a stride of 16. We also use a 1-GPU implementation, and the effective mini-batch size is 2 images by setting the $iter\_size$ to 2. The whole network is trained for 80k iterations with a learning rate of 0.001 and then for 30k iterations with 0.0001. In addition, the context prior is proposed to further boost the performance while keeping the iterations unchanged. Finally, we also perform multi-scale training with the shorter sides of images are randomly resized from 480 to 864.

Table~\ref{comparions with Faster and RFCN} shows the detailed comparisons with Faster R-CNN and R-FCN. As we can see that our single model achieves a mAP of $81.7\%$, which outperforms the R-FCN by 2.2 points. However, while embedding the context prior to the global branch, our mAP rises up to $82.1\%$, which is the current best single model detector to our knowledge. Moreover, we also evaluate the inference time of our network using a NVIDIA TITAN X GPU (pascal) along with CUDA 8.0 and cuDNN-v5.1. As shown in the last column of Table~\ref{comparions with Faster and RFCN}, our method is slightly slower than R-FCN, which also reaches a real-time speed (\ie 8.2 fps or 9.8 fps without context) and achieves the best trade-off between accuracy and speed. We argue that the sharing process of feature extraction between two branches and the design of lightweight RoI-wise subnetwork after RoI pooling both greatly reduce the model complexity.

As shown in Table~\ref{comparisons with state-of-the-art}, we also compared our method with other state-of-the-art single model. We found that our method outperforms the others with a large margin, including the advanced end-to-end SSD method~\cite{liu2016ssd}, which requires complicated data augmentation and careful training skills. Just as discussed earlier, CoupleNet shows a large gain over the classes with occlusions, truncations and considerable background information, like sofa, person, table and chair, which verifies our analyses. We also observed a large improvement for airplane, bird, boat and pottedplant, which usually have class-specific backgrounds, \ie the sky for airplane and bird, water for boat and so on. Therefore, the context surrounding the objects provides an extra auxiliary discrimination.

\subsection{Results on VOC2012}
\begin{figure*}[!thbp]
\begin{center}
\subfigure{
\includegraphics[width=0.2\linewidth]{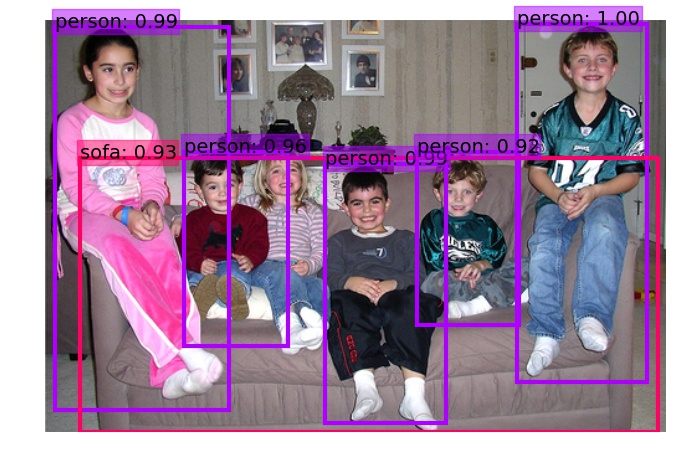}}\
\subfigure{
\includegraphics[width=0.2\linewidth]{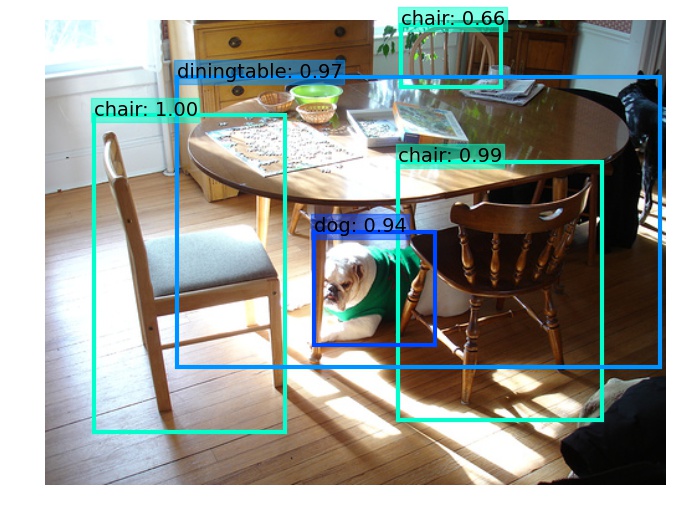}}\
\subfigure{
\includegraphics[width=0.2\linewidth]{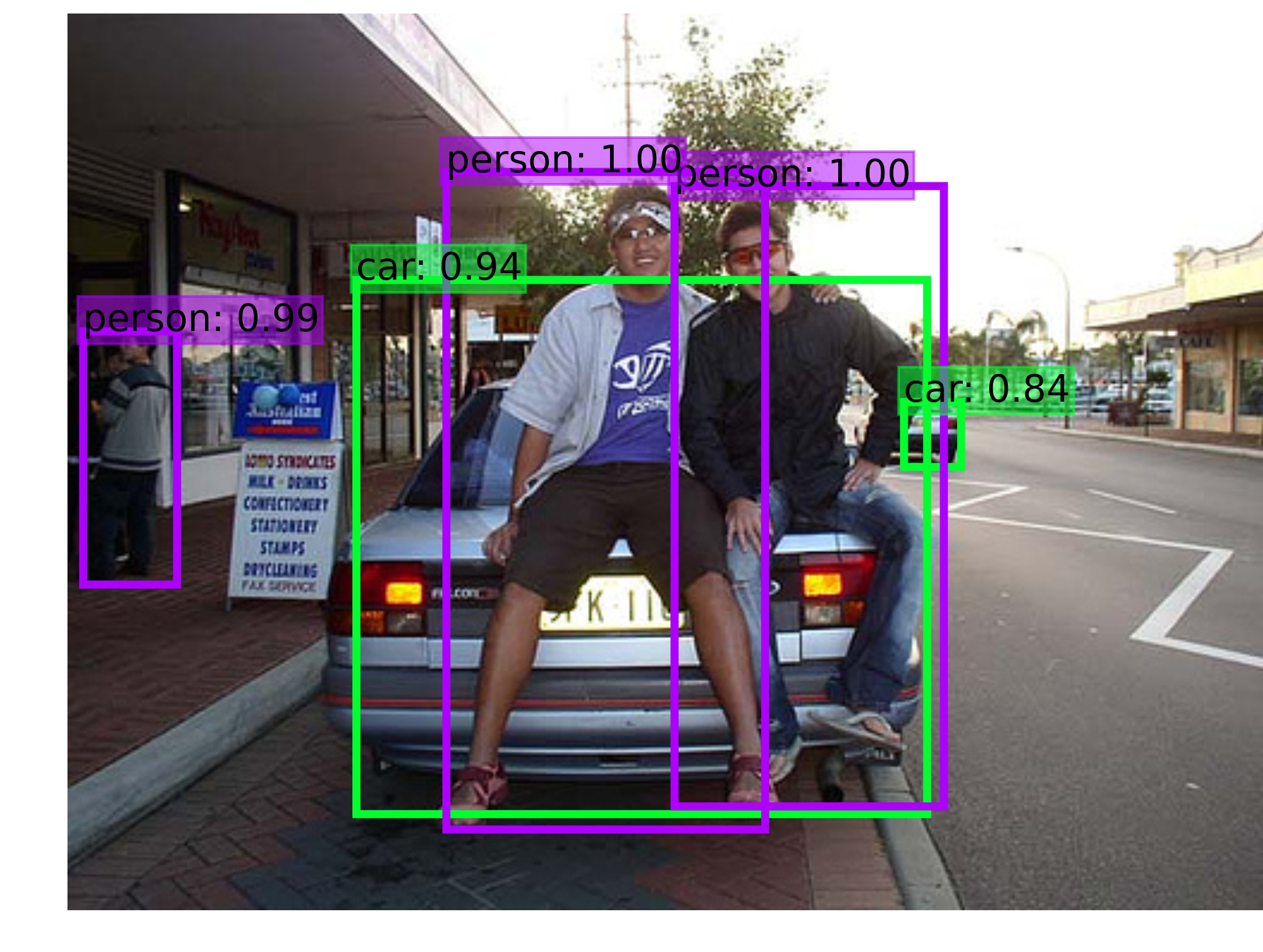}}\
\subfigure{
\includegraphics[width=0.2\linewidth]{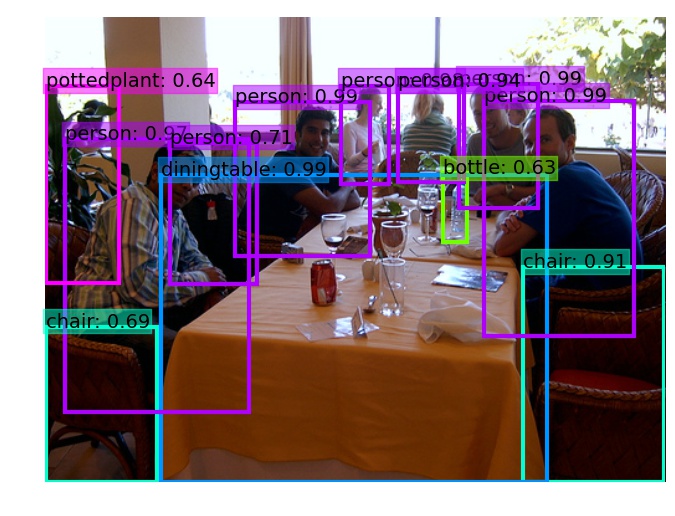}}\\
\vspace{-10pt}
\subfigure{
\includegraphics[width=0.2\linewidth]{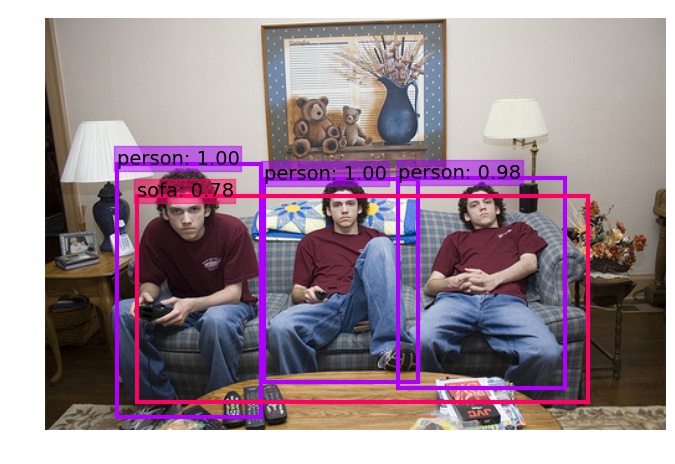}}\
\subfigure{
\includegraphics[width=0.2\linewidth]{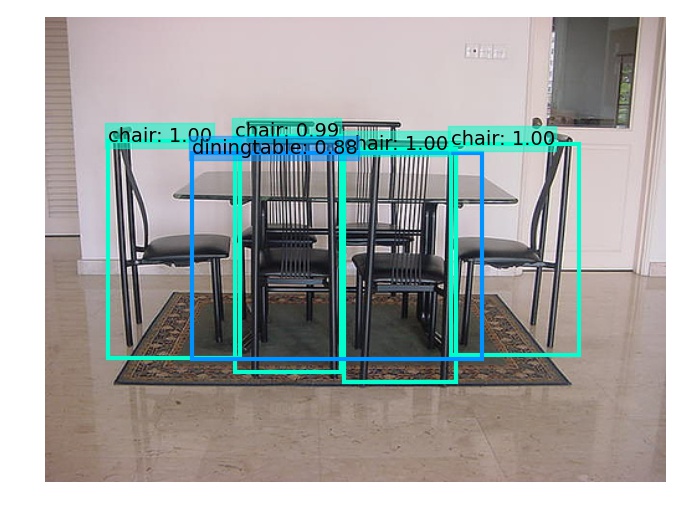}}\
\subfigure{
\includegraphics[width=0.2\linewidth]{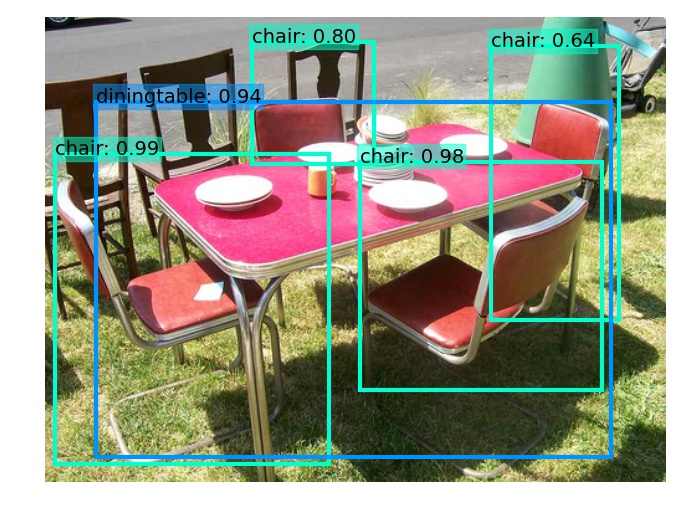}}\
\subfigure{
\includegraphics[width=0.2\linewidth]{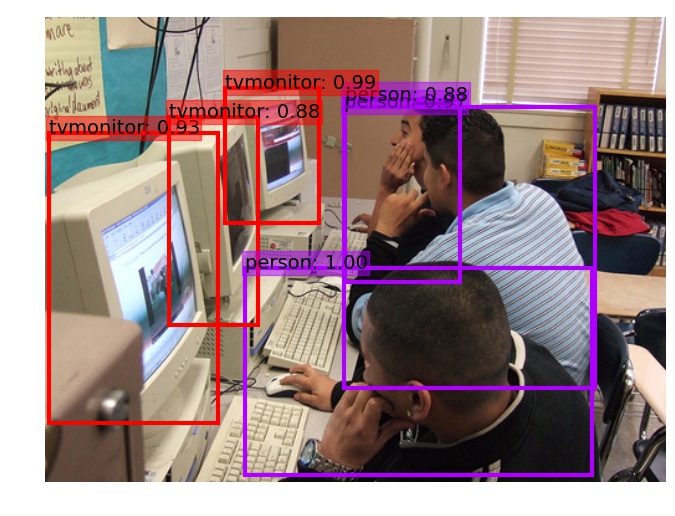}}\\
\vspace{-10pt}
\subfigure{
\includegraphics[width=0.2\linewidth]{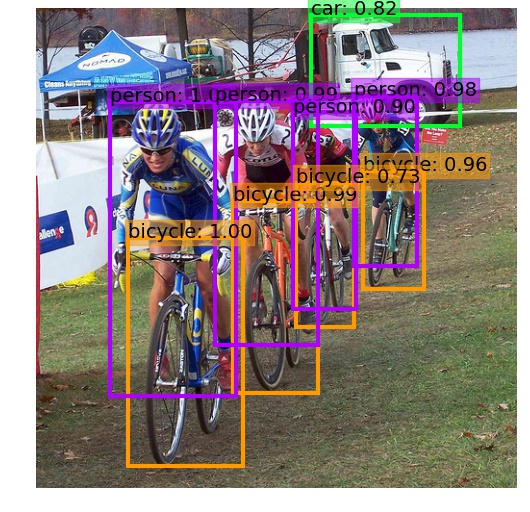}}\
\subfigure{
\includegraphics[width=0.2\linewidth]{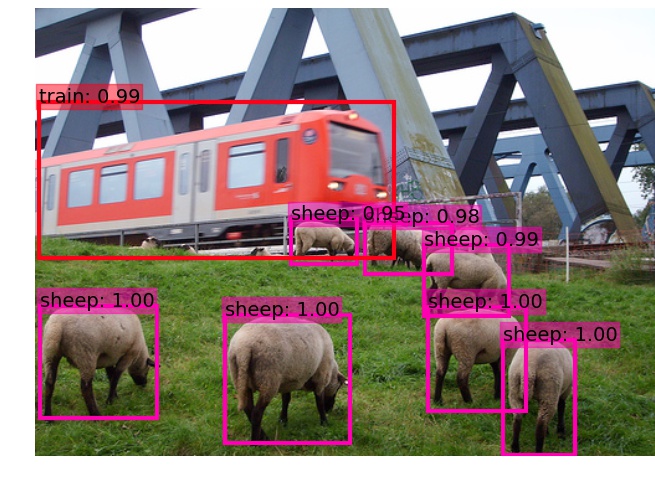}}\
\subfigure{
\includegraphics[width=0.2\linewidth]{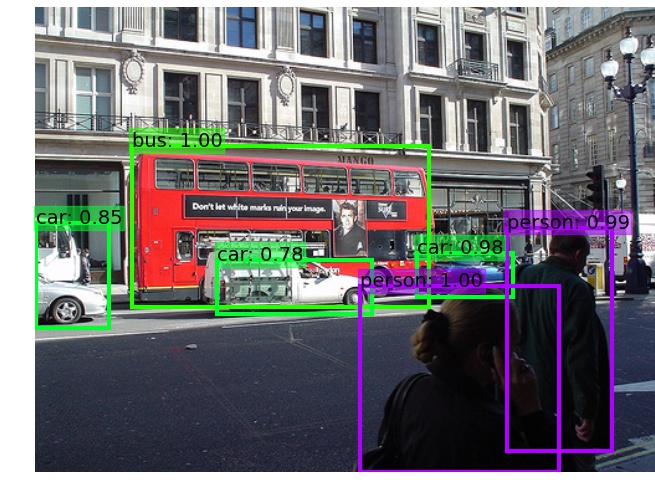}}\
\subfigure{
\includegraphics[width=0.2\linewidth]{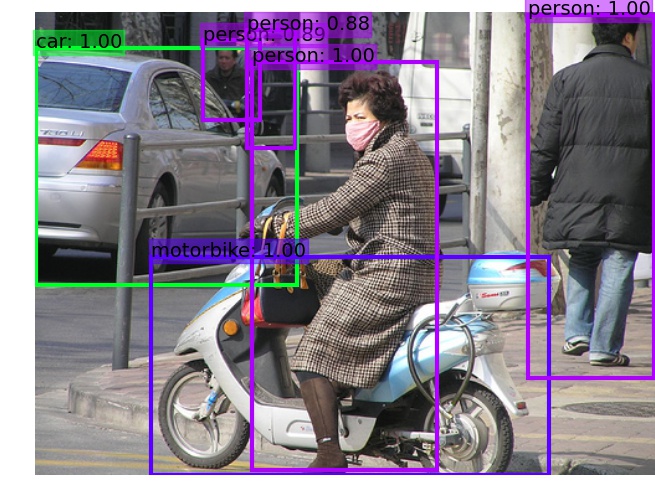}}\\
\vspace{-10pt}
\subfigure{
\includegraphics[width=0.2\linewidth]{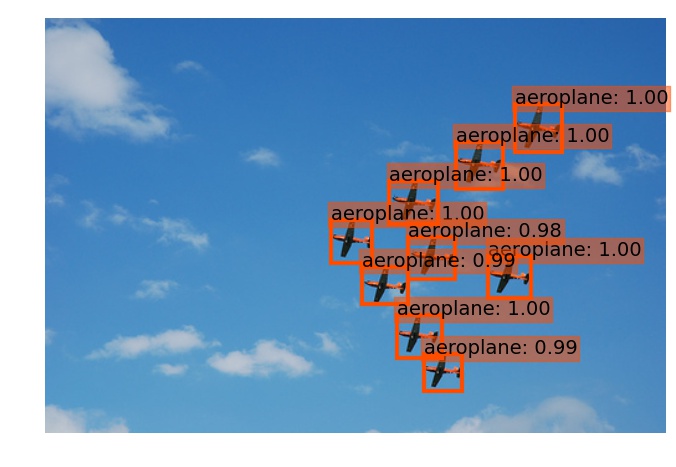}}\
\subfigure{
\includegraphics[width=0.2\linewidth]{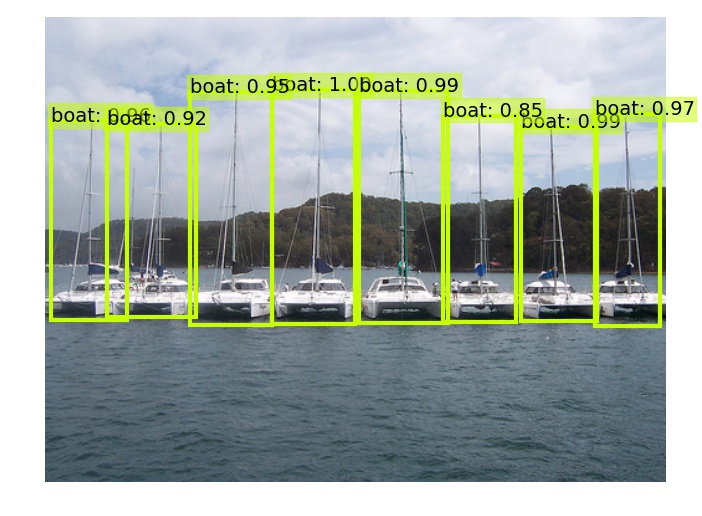}}\
\subfigure{
\includegraphics[width=0.2\linewidth]{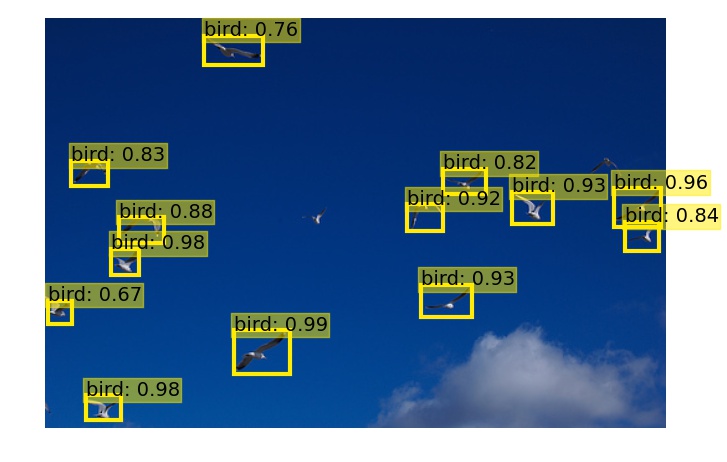}}\
\subfigure{
\includegraphics[width=0.2\linewidth]{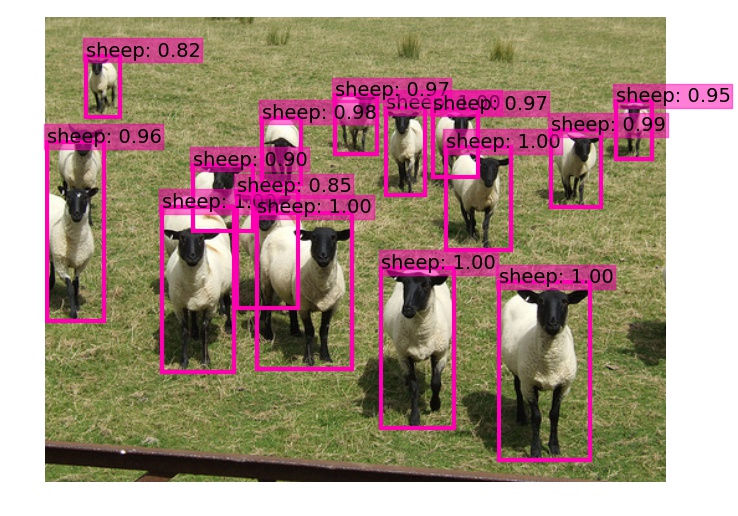}}\\
\end{center}
  \caption{\textbf{Detection examples of CoupleNet on PASCAL VOC 2012 test set.} The model was trained on the union of VOC07 trainval+test and VOC12 trainval ($80.4\%$ mAp). Our method works well with the occlusions, truncations, inter-class interference and clustered background. CoupleNet also shows good performance for the categories with class-specific backgrounds, \eg airplane, bird, boat, \etc. A score threshold of 0.6 is used to draw the detection bounding boxes. Each color is related to an object category.}
\label{voc12results}
\end{figure*}

We also evaluate our method on the more challenging VOC2012 dataset by submitting results to the public evaluation server. We use VOC07 trainval, VOC07 test and VOC12 trainval as the training set, which consists of 21k images in total. We also follow the similar hyper-parameter settings in VOC07 but change the iterations, since there are more training images. We train our models with 4 GPUs, and the effective mini-batch size thus becomes 4 (1 per GPU). As a result, the network is trained for 60k iterations with a learning rate of 0.001 and 0.0001 for the following 20k iterations. Table~\ref{VOC12} shows the results on the VOC2012 test set. Our method obtains a top mAP of $80.4\%$, which is 2.8 points higher than R-FCN. We note that without using the extra tricks in the testing phase, our detector is the first one with a mAP higher than $80\%$. Similar promotions over the specific classes analysed in VOC07 are also observed, which once again validates the effectiveness of our method. Figure~\ref{voc12results} shows some detection examples on VOC 2012 test set.

\subsection{Results on MS COCO}
Next we present more results on the Microsoft COCO object detection dataset. The dataset consists of 80k training set, 40k validation set and 20k test-dev set, which involves 80 object categories. All our models are trained on the union set of 80k training set and 40k validation set, and evaluated on 20k test-dev set. The COCO standard metric denotes as AP, which is evaluated at $IoU \in [0.5:0.05:0.95]$. Following the VOC2012, a 4-GPU implementation is used to accelerate the training process. We use an initial learning rate of 0.001 for the first 510k iterations and 0.0001 for the next 70k iterations. In addition, we conduct multi-scale training with the scales are randomly sampled from $\{480,576,672,768,864\}$ while testing in a single scale.

Table~\ref{COCO} shows our results. Our single-scale trained detector has already achieved a result of $33.1\%$, which outperforms the R-FCN by 3.9 points. 
In addition, the multi-scale training further improves the performance up to $34.4\%$. Interestingly, we observed that the more challenging the dataset, the more the promotion (\eg, $2.2\%$ for VOC07, $2.8\%$ for VOC12 and $4.5\%$ for COCO, all in multi-scale training), which directly proves that our approach can effectively cope with a variety of complex situations.

\section{Conclusion}
In this paper, we present the CoupleNet, a concise yet effective network that simultaneously couples global, local and context cues for accurate object detection.
Our system naturally combines the advantages of different region-based approaches with the coupling structure. With the combination of local part representation, global structural information and the contextual assistance, our CoupleNet achieves state-of-the-art results on the challenging PASCAL VOC and COCO datasets without using any extra tricks in the testing phase, which validates the effectiveness of our method.

{\small
\bibliographystyle{ieee}
\bibliography{CoupleNet}
}
\end{document}